\documentclass[letterpaper, 10 pt, conference]{ieeeconf}
\IEEEoverridecommandlockouts                              % This command is only needed if 
\usepackage{amsmath,bm,amsfonts,cases}
\usepackage{shortcuts,subfiles,balance}
\usepackage{booktabs,xcolor}
\usepackage{graphicx}
\usepackage{float}
\usepackage{url}
\usepackage{optidef}    % optimization problems
\usepackage{mathrsfs}
\usepackage[hidelinks]{hyperref} % for urls
\usepackage[format=plain,font=footnotesize,labelsep=colon]{caption}

\title{\LARGE \bf Chasing Stability: Humanoid Running via Control Lyapunov Function Guided Reinforcement Learning}
\author{Zachary Olkin, Kejun Li, William D. Compton, Aaron D. Ames$^{}$% <-this % stops a space
\thanks{This research is supported by Technology Innovation Institute (TII) and the National Science Foundation Graduate Research Fellowship.}
\thanks{The supplementary video can be found here: \url{https://youtu.be/zCtDQuZAomI}}
}

\begin{document}
\maketitle

\begin{abstract}
Achieving highly dynamic behaviors on humanoid robots, such as running, requires controllers that are both robust and precise, and hence difficult to design. Classical control theory offers principled tools for stabilizing such systems but struggles to synthesize real-time controllers for their nonlinear, hybrid dynamics; reinforcement learning (RL) handles these dynamics well but typically depends on extensive hand-tuning of heuristic rewards. In this work, we embed ideas from nonlinear control theory, specifically control Lyapunov functions (CLFs), along with optimized reference trajectories for the hybrid system into the reinforcement learning reward. This approach, CLF-RL, replaces the need to handcraft and tune heuristic reward terms with easy to tune stability informed rewards. We further develop a trajectory optimization routine that can handle the hybrid structure of the system to produce dynamically feasible, multi-domain, trackable references; using full-order dynamics rather than a kinematic approximation yields up to a 1.7x improvement in position tracking. The resulting policy runs on a Unitree G1 humanoid with a full flight phase, operating reliably in tight corridors such as a treadmill as well as in outdoor environments, all while remaining robust to disturbances applied to the torso and feet. Moreover, it achieves accurate global reference tracking utilizing only on-board sensors, making a critical step toward integrating these dynamic motions into a full autonomy stack. 

\end{abstract}

\section{Introduction}
Humanoid running is a challenging task that involves executing highly dynamic motion on a nonlinear and hybrid system. Achieving performant and robust running demands controllers that can reject disturbances arising from model mismatch and environmental uncertainty, all while operating near the limits of the robot’s dynamic capabilities. Running inherently involves alternating between a flight phase, where both feet are off the ground, and a single-support phase, where only one foot is in contact. Effectively handling control across these hybrid domains is critical, as improper treatment can lead to instability.

Bipedal running has been studied for decades with early examples including the Raibert heuristic \cite{raibert1986legged}. In the 2010's, a number of planar bipeds were developed and running was achieved \cite{ma2017bipedal,sreenath_embedding_2013,morris_achieving_2006}. These methods fall under the category of Hybrid Zero Dynamics (HZD) \cite{westervelt2003hybrid} where offline trajectory optimization leveraging the idea of virtual constraints is used to generate a stable and periodic trajectory. Then, online, the trajectories are tracked using tools from nonlinear control theory such as feedback linearization and control Lyapunov functions (CLFs). These controllers have been shown to be certifiably stable if the convergence to the virtual constraints is sufficiently quick relative to the destabilizing effect of the foot-ground impact \cite{ames2014rapidly}. In general, these methods optimize for a steady-state motion (i.e. a periodic orbit) while the ability to get to the steady state motion is entirely dependent on the region of attraction of the tracking controller. Because these controllers operate only on the continuous dynamics, their capacity to generate transient and robust behaviors is inherently limited.

\begin{figure}[t!]
    \centering
    \includegraphics[width=1.0\linewidth]{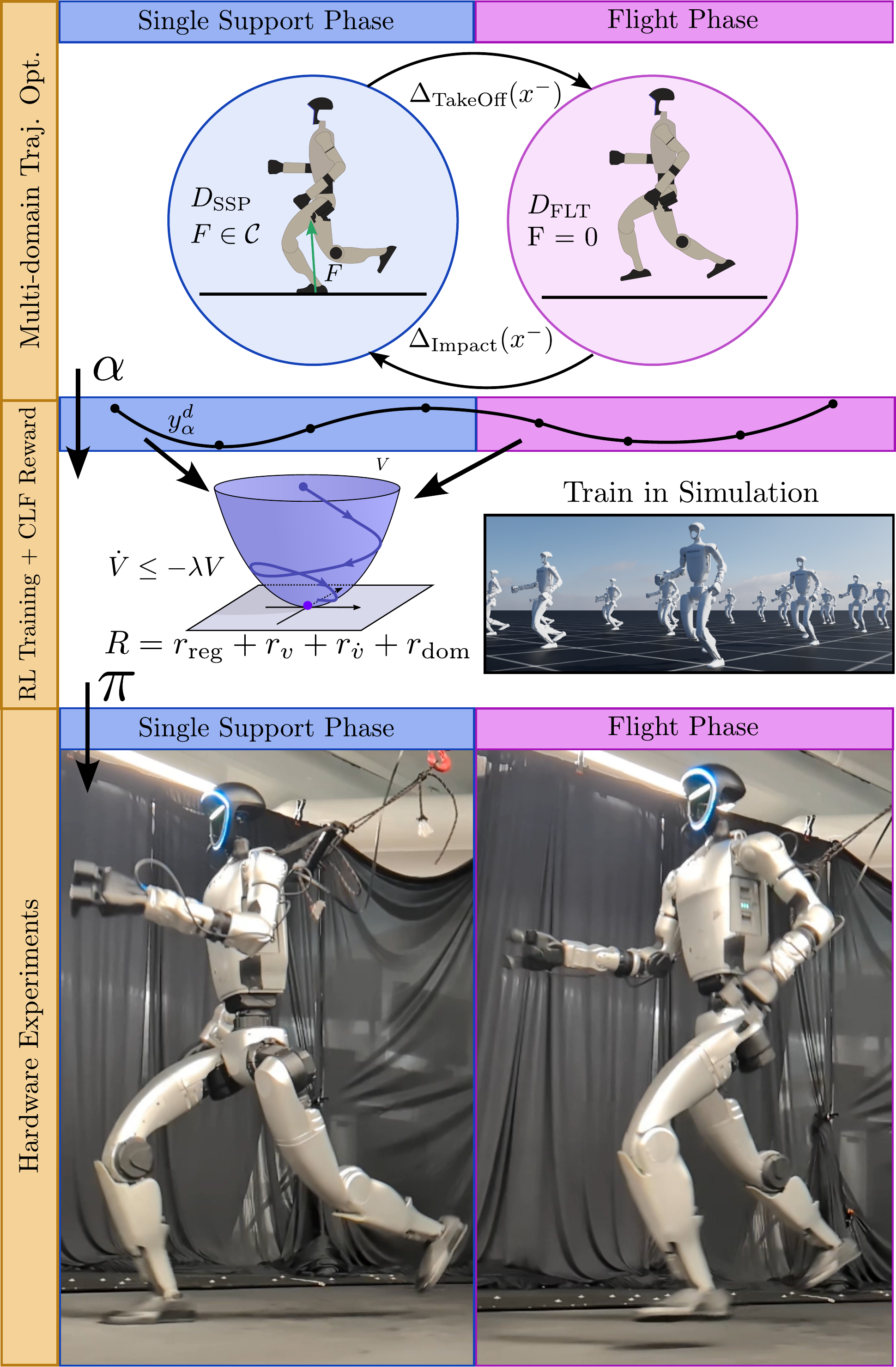}
    \caption{Overview of our approach. Trajectory optimization through each of the hybrid domains generates desired trajectories, which are used to construct a CLF-based reward. An RL policy is trained in simulation with this reward and deployed on a real humanoid robot.}
    \vspace{-6mm}
    \label{fig:hero_fig}
\end{figure}

One of the difficulties with generating transient motions lies in the ability to reason through contact, specifically determining the subsequent contact schedule. A number of model predictive control (MPC) schemes have attempted to solve this issue through various numerical methods. Contact implicit MPC (CI-MPC) uses gradient-based optimization to implicitly yield a hybrid domain sequence, but has not yielded bipedal running and is quite computationally intensive in general \cite{kurtz_inverse_2023,cleach_fast_2023,wensing_optimization-based_2022}. Sample-based methods circumvent the need for an explicit domain sequence by rolling out sampled inputs and optimizing the inputs in an MPC fashion \cite{xue_full-order_2024,williams_model_2017,howell_predictive_2022}. Yet these methods still require large amounts of on-board compute and have not yet produced humanoid running.

Reinforcement learning (RL) has recently emerged as a dominant approach for controlling legged robots \cite{zhuang_humanoid_2024,radosavovic_real-world_2024,wang_beamdojo_2025,hoeller_anymal_2024,lee_integrating_2024}. RL methods are attractive due to their robustness, ability to generate diverse motions, lightweight on-board execution requirements, and their capacity to learn contact-rich behaviors directly from experience rather than requiring explicit contact modeling. Notably, RL has achieved bipedal running \cite{crowley2023optimizing,li_reinforcement_2024}. However, these RL pipelines require extensive hand tuning of rewards to produce a performant and robust policy. Poorly shaped rewards can lead to unstable learning, failure to achieve the desired behaviors, or prohibitively long training times. To mitigate these issues, a number of works have merged pre-computed trajectories with RL. This includes reduced-order model trajectories \cite{batke_optimizing_2022,green_learning_2021} and full-order model trajectories such as from the HZD framework  \cite{li_reinforcement_2021}. Yet, even in these cases, humanoid running is lacking in the literature, and the rewards generally incentivize being close to the trajectories in an ad-hoc manner. Building on the ideas of \cite{li2025clf}, we embed a CLF tracking controller's stability condition and Lyapunov function into the reward and additionally generalize to arbitrary hybrid systems by adding a simple contact driven reward to extend the method.

Many prior works focus on bipeds, not humanoids, and have not shown running with low positional drift required for treadmill operation. Hand designed rewards are often used and the policies may fail to match the desired velocity \cite{li_reinforcement_2024}. Even when fast locomotion is achieved, a flight phase is not necessarily attained \cite{batke_optimizing_2022}. Alternative approaches use RL to imitate human motion data \cite{he_asap_2025,liao_beyondmimic_2025,peng_deepmimic_2018,xie_kungfubot_2025}. Although such methods can reproduce the demonstrated motion, they inherently rely on human data and have not shown the ability to produce steady-state motions, like running, with tracking capabilities ready for use in an autonomy stack. In contrast, our proposed method provides a principled way of synthesizing running controllers for full humanoids, not just bipeds, with minimal tuning and without relying on human data. The resulting policy produces both transient and steady state running motions with accurate position and velocity tracking.

The contributions of this paper are as follows. (1) To the best of our knowledge this is the first demonstration of multi-domain (flight and ground phase) humanoid running through reference tracking RL without using human data or animation. To accomplish this, we propose a general trajectory optimization method utilizing full order dynamics through a hybrid graph that is solvable in minutes and can yield trackable dynamic trajectories. (2) Although it is easier to use kinematic trajectories, we show that dynamics is important in the optimization for motions like running. We compare tracking kinematically optimized solutions with dynamically optimized solutions and show that having the dynamics improves both tracking performance and training metrics consistently. (3) We demonstrate this controller on a Unitree G1 humanoid both indoors and outdoors. The policy is controllable enough for straight line running in tight corridors, such as on a treadmill, or outside, as demonstrated using a layered control architecture. Further, the policy is robust to disturbance forces and uneven ground. Fig. \ref{fig:hero_fig} shows an overview of the proposed framework.

\section{Preliminaries}
\label{sec:prelim}
\subsection{Hybrid Systems}
Legged locomotion can be represented as a nonlinear hybrid dynamical system with impulse effects, including both continuous and discrete states. In the case of a humanoid, the system naturally involves multiple domains. We denote the system as a 5-tuple: $\mathscr{H} = (\mathcal{D}, \mathcal{S}, \Gamma, \Delta, \mathcal{F})$ where the domain $\mathcal{D} := \mathcal{X} \times \mathcal{U}$ consists of the state manifold and input space, $\mathcal{S}$ denotes the set of guards, $\Gamma$ the set of transitions, $\Delta$ are the reset-maps, and $\mathcal{F}$ the continuous dynamics. 
A transition from domain $D_i$ to $D_{i+1}$ is denoted as $\gamma := (D_i, D_{i+1}) \in \Gamma$. Then let $\mathcal{S}_{\gamma} \subset \mathcal{X}$ and $\Delta_\gamma$ denote the guard and associated reset map for a given transition. Similarly, let $f_{D} \in \mathcal{F}$ denote the dynamics for a given domain $D \subset \mathcal{D}$. In our case, we consider the single support phase (SSP) and the flight phase (FLT) for running; while standing requires the double support phase (DSP).

We can write the hybrid dynamics of the system as 
\begin{numcases}{\mathcal{H} :=}
\dot{x} = f_{D}(x) + g_{D}(x)\,u_{D} & $x \in {D} \setminus \mathcal{S}_\gamma$, \label{eq: continuous_dynamics}
\\
x^+ = \Delta_{\gamma}(x^-) & $x^- \in S_\gamma$, \label{eq: discretecontrol}
\end{numcases}
where $x \in \mathcal{X}$, $u \in \mathcal{U}$. For more details on hybrid modeling for bipeds, refer to \cite{grizzle_3d_2010}.

The continuous dynamics can be derived as:
\begin{align}
  M(q)\ddot{q} + H(q,\dot{q}) = B u + J_h(q)^TF \label{eq:dynamics} \\
  J_h(q) \ddot{q} + \dot{J}_h(q,\dot{q})\dot{q} = 0 \label{eq:hol_dynamics}
\end{align}
where $M(q):\mathcal{Q}\to \mathbb{R}^{n\times n}$ is the mass-inertia matrix, $H: \mathcal{Q} \times T\mathcal{Q} \to \mathbb{R}^n$ contains the Coriolis and gravity terms, and $B\in\mathbb{R}^{n\times m}$ is the actuation matrix. The Jacobian of the holonomic contact constraint is $J_h(q) \in \mathbb{R}^{h \times n}$ and the associated constraint wrench is $F \in \mathbb{R}^h$. For the patch contact used here $h = 6$. Then the state is $x = [q, v]^T$.

We consider the following transitions: $(\text{SSP}, \text{FLT})$ (take off) and $(\text{FLT}, \text{SSP})$ (impact) with the associated reset maps:
\begin{equation}
    \Delta_{(\text{SSP}, \text{FLT})}(x) = Ix.
\end{equation}
and
\begin{equation}
    \Delta_{(\text{FLT}, \text{SSP})}(x) : \begin{bmatrix} q \\ \dot{q} \end{bmatrix} \rightarrow \begin{bmatrix}
        q \\ 
        (I - M^{-1}J_h^T(J_hM^{-1}J_h^T)^{-1}J_h)\dot{q}
    \end{bmatrix} 
    \label{eq:impact_map} 
\end{equation}
with $I$ the identity matrix.

The domains are defined by the robot’s contacts with the ground, while the guards are determined by the ground height. Each domain specifies the number of active contact forces and their associated Jacobians. In the flight phase, no ground contacts exist, so the ground reaction force is zero throughout the domain ($F = 0$) and there is no holonomic constraint. In the single support phase (SSP), the contact force must lie within the friction cone $\mathcal{C}$, and a holonomic no-slip constraint is imposed at the contact point.

\subsection{Hybrid Zero Dynamics}
Hybrid Zero Dynamics (HZD) uses offline trajectory optimization to generate trajectories that lie on a stable zero dynamics manifold, $\mathcal{Z}^\alpha$. This zero dynamics manifold contains the dynamics of the system after the virtual constraints, $y_\alpha$, are satisfied. These virtual constraints are end-effector or joint trajectories with some parameterization, $\alpha$ (typically Bézier curve coefficients). To achieve certifiable stability, HZD designs the gait as a steady-state periodic orbit in state space, allowing tools from control theory to be used for analysis of the orbit. Owing to this periodicity, only one leg’s gait needs to be explicitly designed, as the other can be obtained through a symmetry mapping. The ideas of virtual constraints and periodic orbits will be leveraged in the multi-domain trajectory optimization later.

%When underactuated, we cannot define outputs on every degrees on freedom, consequently, when these outputs are tracked, a residual set of dynamics, known as zero dynamics, is left. 

The key requirement for periodicity in a hybrid system is impact invariance, enforced in the optimization as:
\begin{align}\label{eq:impact_invariance}
    \Delta_{\gamma}(\mathcal{Z}^{\alpha} \cap \mathcal{S}_{\gamma}) \subset \mathcal{Z}^{\alpha}.
\end{align}

\subsection{Control Lyapunov Functions}
Control Lyapunov Functions (CLFs) are a classic tool in nonlinear control theory for designing certifiably stable controllers. They are the natural extension of Lyapunov functions for systems with an input. For a smooth nonlinear control-affine system of the form \eqref{eq: continuous_dynamics}, a continuously differentiable, positive definite function $V(x) : \mathcal{X} \rightarrow \mathbb{R}$ is called an exponentially stabilizing CLF if the set
\begin{align}
    \{u \: | \nabla_x V(x)(f(x) + g(x)u)  < -\lambda V(x)\}
    \label{eq:clf_condition}
\end{align}
is non-empty for scalar $\lambda > 0$. This inequality guarantees that $V(x)$ decreases along system trajectories, implying exponential convergence of the state $x$ to the origin (or a desired manifold).

For hybrid systems such as humanoid locomotion with discrete impact events, CLFs can still be applied within each continuous domain of the dynamics. Prior work has shown that if the rate of convergence in the continuous phase is sufficiently fast, it can offset the destabilizing effects of impacts \cite{ames2014rapidly}, providing a principled way to certify stability despite hybrid transitions.

\subsection{Reinforcement Learning}
% Do I even want this section?
Reinforcement learning (RL) is a powerful framework for synthesizing robust, high-performing control policies through training in simulation. At its core, RL seeks to solve the stochastic optimal control problem:
\begin{equation}
    \pi^* = \arg \max_{\pi} J(\pi)
\end{equation}
where $\pi$ is the control policy and $J$ is the expected reward: $J := \mathbb{E}[R]$ for a reward, $R$. A critical aspect of RL is therefore the design of $R$ to induce the desired behavior. With poorly shaped rewards, training can be unstable or the resulting policy may not yield the desired result. However, even with well-designed rewards, policies trained in simulation may fail to transfer to real systems due to model mismatch and unmodeled dynamics. To address this challenge, domain randomization is often used to robustify the policy and facilitate sim-to-real transfer, thus preventing over exploitation of the simulation dynamics. 

% Commonly domain randomization is used to robustify the policy and facilitate \textit{sim2real} transfer. 

%Through extensive offline training utilizing CLF-guided rewards and domain randomization, we can generate policies that allow our robot to run.

\begin{figure} 
    \vspace{2mm}
    \centering
    \includegraphics[width=1.0\linewidth]{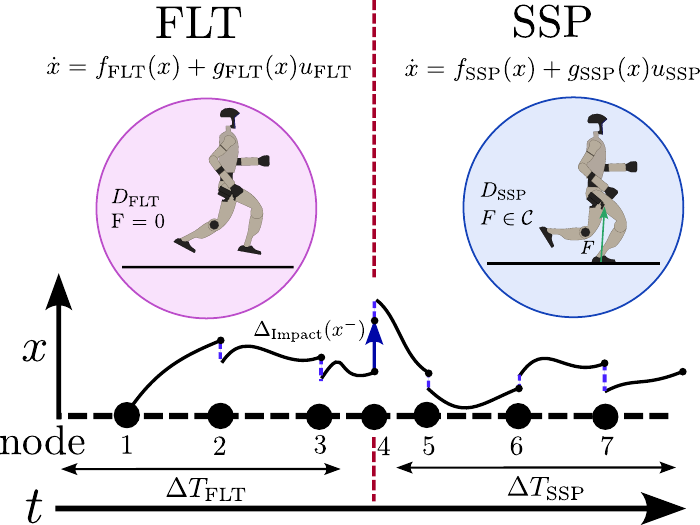}
    \caption{Visual depiction of the multiple shooting trajectory optimization problem. Two hybrid domains are shown and the node where they meet utilizes the associated reset map. Within each domain, a number of optimization nodes are used with variable times associated with each domain.}
    \vspace{-5mm}
    \label{fig:hybrid_opt}
\end{figure}

\section{Running Controller}
\label{sec:methods}
\subsection{Multi-Domain Trajectory Optimization}
A gait library spanning speeds from 1.1 m/s to 3.0 m/s is generated using a multi-domain trajectory optimization routine. In addition, a standing pose is also generated. The hybrid system is modeled with three domains: single support phase (SSP), double support phase (DSP), and flight phase (FLT), each requiring its own dynamics and constraints. To correctly apply the corresponding constraints, it is essential to know which optimization nodes belong to each domain. Therefore, the domain sequence is fixed, while the timing within each domain is left to be optimized. This is achieved by using a variable time assigned to each domain and making the integration time for each node an even division of this total time. The resulting problem is formulated as a multiple-shooting trajectory optimization, as illustrated in Fig. \ref{fig:hybrid_opt}. This generalizes the trajectory optimization routine in \cite{li2025clf} to handle multi-domain hybrid systems. The code is setup to take in an arbitrary hybrid graph with arbitrary edges, thus making the proposed optimization problem very general.

To generate periodic motions, we optimize for a steady-state gait by enforcing a periodicity constraint. Because only half of the gait is optimized, rather than a full gait cycle, the states must be symmetrically re-mapped at the end of the trajectory, and the impact map \eqref{eq:impact_map} must be applied to transition between steps. The running gaits are therefore constructed using a SSP followed by a FLT, with a takeoff condition between them and both the impact map and leg re-mapping applied at the end. In contrast, the standing pose is generated using only the DSP.

To enforce the domain sequence, we create a set of node indexes $\mathcal{J}$ at which the domain must switch. Associated with each of these indexes are a corresponding guard and reset map. We denote the set of these guards and reset maps as $S_\mathcal{J}$ and $\Delta_\mathcal{J}$. The optimization only operates on discretized dynamics, which we denote as $\bar{f}_D$ and $\bar{g}_D$. This yields the full optimization problem as follows:
\begin{mini!}[1]
{\alpha, x, u}{\Phi(x, u)}
{\label{eq:hybrid-opti}}{}
\addConstraint{x_{k + 1}}{= \bar{f}_{D_k}(x_k) + \bar{g}_{D_k}(x_k)u_k \label{eq:opti-dynamics}}{\quad k = 0...N}
\addConstraint{x_{k}}{\in S_{k}}{\quad k \in \mathcal{J}}
\addConstraint{x_{k + 1}}{= \Delta_{k}(x_{k - 1})}{\quad k \in \mathcal{J}}
\addConstraint{\Delta_{k}(S_{k}}{\cap \mathcal{Z}^\alpha) \subset \mathcal{Z}^\alpha}{\quad k \in \mathcal{J}}
\addConstraint{y^d_\alpha(t_k)}{= y(x_k)}{\quad k = 0...N}
\addConstraint{x_{\text{min}}}{\leq c_x(x_k) \leq x_{\text{max}}}{\quad k = 0...N}
\addConstraint{u_{\text{min}}}{\leq c_u(u_k) \leq u_{\text{max}}}{\quad k = 0...N}
\end{mini!}

To solve the optimization problem, Casadi \cite{andersson_casadi_2019}, Pinocchio \cite{carpentier2019pinocchio}, and IPOPT \cite{wachter_implementation_2006} are used. Given a total gait time, we can enforce the average speed by fixing a forward step length as a constraint. Once the fastest gait is generated, we can use it to warm start the gait library creation. Specifically, solutions for the faster gaits are used to warm start future solves of slower gaits, which contain modified step length constraints and arm swing amounts, while keeping the total time constant. This allows us to generate incrementally slower gaits very efficiently, and prevents neighboring gaits from converging to drastically different solutions.

By designing trajectories through optimization we can embed hard constraints directly into the problem to generate the desired gait. This approach allows for rapid iteration, usually requiring seconds or minutes to solve for a gait. Virtual constraints are incorporated by enforcing that a Bézier curve fits the optimized nodes of the desired outputs. Outputs include various end effector poses and robot joints. By fitting this Bézier curve, the resulting output could be tracked using a feedback linearization controller \cite{csomay-shanklin_multi-rate_2022} as done in classical HZD. All position-based virtual constraints are relative to the location of the stance foot, not a global position. During the flight phase, the virtual constraints are relative to the last location of the stance foot in the single support phase, as the stance foot is allowed to move in the flight phase.

% We can write the optimization as 
% \begin{align*} 
%    \{\alpha^*,X^*\} = & \argmin_{\alpha,X}    \Phi(X) \\
%     \text{s.t.}\quad 
%     & \dot{x} = f_D(x) + g_D(x)u \; x \in D \setminus S  \tag{Dynamics} \\
%     & x^+ = \Delta_S(x^-) \; x^- \in S \tag{Reset Maps} \\
%     & \Delta(\mathcal{S} \cap \mathcal{Z}^\alpha) \subset \mathcal{Z}^\alpha \tag{Periodicity} \\
%     & X_{\text{min}}  \preceq X \preceq X_{\text{max}} \tag{Decision Variables} \\
%     & c_{\text{min}}  \preceq c(X) \preceq c_{\text{max}} \tag{Physical Constraints},
% \end{align*}
% which generates the trajectory, $X^*$ and bezier parameters for the virtual constraints, $\alpha^*$.

\subsection{RL Design}
To enable accurate trajectory tracking, we incorporate control Lyapunov functions (CLFs) into the reward function to promote tracking and convergence to the steady-state trajectories. Both the function itself, $V$, and the decreasing condition are embedded into the reward. Each joint or end effector is modeled using a double-integrator system, which allows the use of a quadratic CLF \cite{li2025clf}. 
% Since these are fully actuated outputs, they can, in principle, be feedback linearized to follow arbitrary linear dynamics, such as a double integrator, thereby justifying this reduced-order modeling choice.
\begin{figure}
    \centering
    \includegraphics[width=1.0\linewidth]{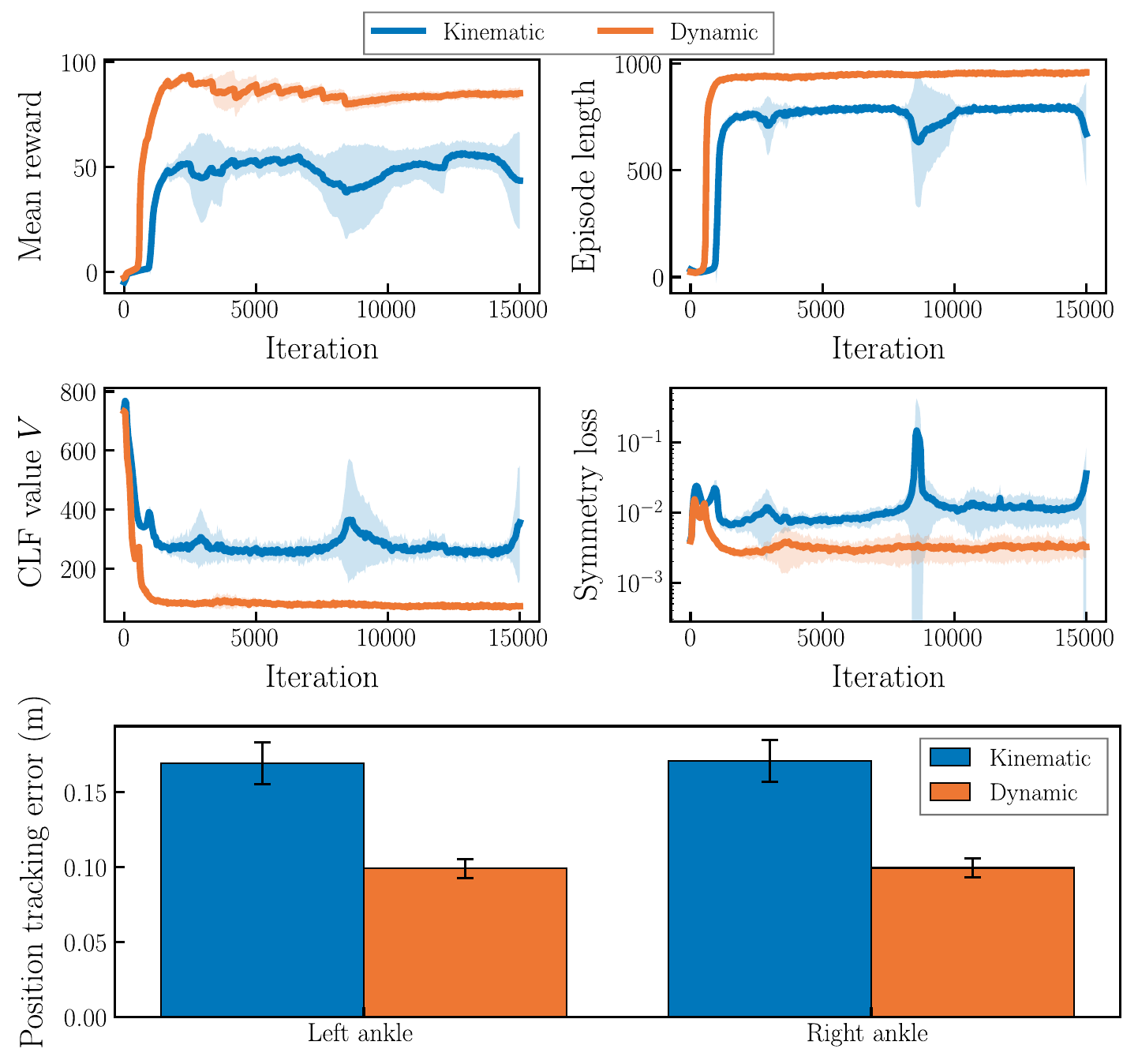}
    \caption{Comparison between kinematic and dynamic trajectory optimization across 5 different seeds tracking a 3.3 m/s trajectory. The shaded region in the training plots and the bars in the tracking plots all show 1 standard deviation. The dynamic shows definitively better training and final tracking performance.}
    \label{fig:kinematic_comparison}
    \vspace{-3mm}
\end{figure}
\begin{figure*}
    \centering
    \includegraphics[width=1.0\linewidth]{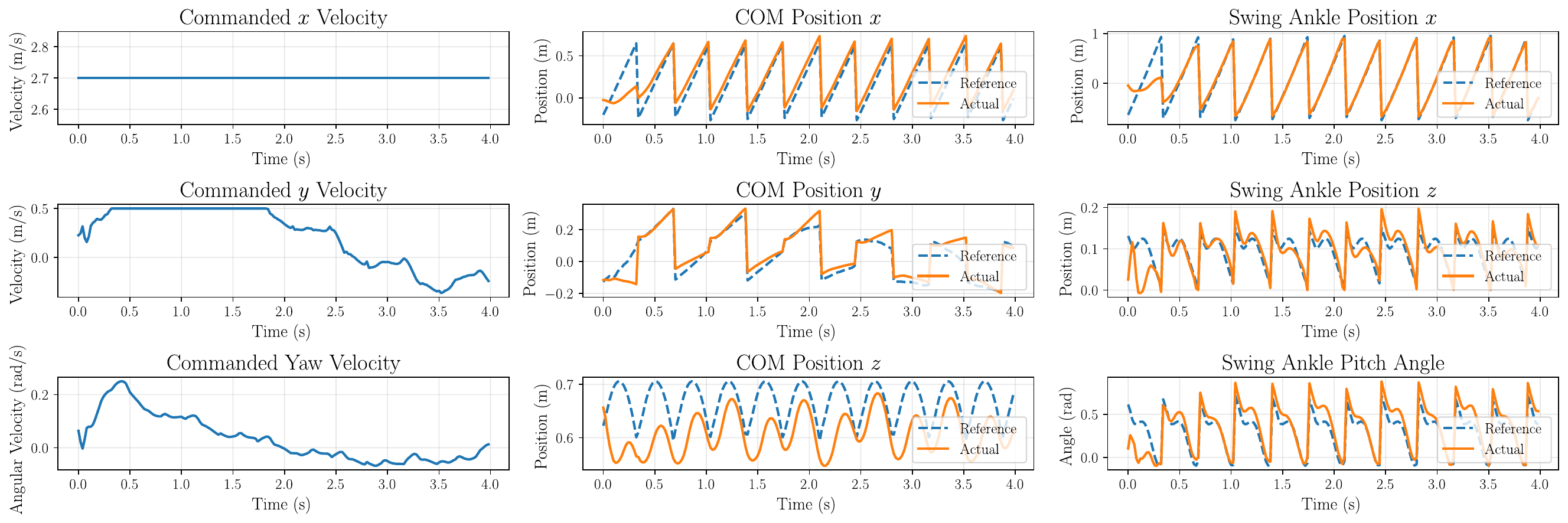}
    \caption{Trajectory tracking of the learned policy for selected virtual constraints in IsaacSim. The policy successfully tracks the reference trajectories and is able to produce the necessary transient response to converge to steady-state motions despite the absence of transient information in the gait library.}
    \label{fig:isaac-traj-tracking}
    \vspace{-5mm}
\end{figure*}
Denote the error between the outputs and their virtual constraints as $\eta$:
\begin{equation}
    \eta(t) = 
    \begin{pmatrix}
    y_\alpha^d(t) - y(q) \\
    \dot{y}_\alpha^d(t) - \dot{y}(q, \dot{q})
    \end{pmatrix}
\end{equation}
then we can write the CLF as 
\begin{equation}
    V(\eta) = \eta^T P \eta
\end{equation}
and further define $V_t := V(\eta(t))$ for ease of notation.
We define a CLF tracking reward:
\begin{align}
r_{v} &= w_{v} \exp\left(\frac{-V_t}{\sigma_{v}}\right).
\label{eq:clf_tracking}
\end{align}
The value of the CLF lies in its stability condition \eqref{eq:clf_condition}, which tells the system to always be converging to the desired point/trajectory. To get this value we require a derivative of $V$, but rather than computing the derivatives analytically we opt for a finite-difference approximation. Then the CLF decay condition reward is:
\begin{equation}
r_{\dot{v}} = -w_{\dot{v}} \;\operatorname{clip}\!\left(\tfrac{\dot{V}_t + \lambda V_t}{\sigma_{\dot{v}}},\,0,\,1\right) 
\label{eq:clf_decrease},
\end{equation}
which penalizes the policy for not converging as fast as \eqref{eq:clf_condition}.

In addition to the CLF terms, we include a reward that encourages the contact sequence underlying the domain of the hybrid system. In the single-support phase, the reward encourages the stance foot to remain stationary, while in the flight phase, contact forces are penalized. The appropriate reward is selected based on the elapsed time within the current gait cycle. We can break the domain reward, $r_{\mathrm{dom}}$ into two parts: $r_{\mathrm{dom}} = \mathbf{1}_{\text{SPP}}r_{\mathrm{hol}} + \mathbf{1}_{\text{FLT}}r_{\mathrm{con}}$ where $\mathbf{1}_{\text{SSP}} = 1$ if the reference gait is in SSP and $\mathbf{1}_{\text{FLT}} = 1$ if the gait is in the flight phase; both are zero otherwise. Each of those terms is then given by:
\begin{align*}
r_{\mathrm{hol}} = &w_{\mathrm{hpos}} \exp\left(-\frac{\|p_{\mathrm{st}} - p_{\mathrm{st}}^0\|}{\sigma_p}\right) + w_{\mathrm{hvel}} \exp\left(-\frac{\|v_{\mathrm{st}}\|}{\sigma_v}\right) \\
r_{\mathrm{con}} = & -w_{\text{con}} \tanh{\Big(\sum_{j = 1}^{2} F_j / \sigma_{\text{con}}\Big)}
\end{align*}
where $p^0_{\mathrm{st}}$ is the position of the stance foot at the beginning of SSP, and $F_j$ is the force on the $j$th foot. We use a curriculum on the contact penalty to make the learning easier to start and end the curriculum after about 8,300 iterations at the final weight. The policy is trained for 15,000 total iterations.

% \begin{figure}
%     \centering
%     \includegraphics[width=1.0\linewidth]{figures/clf_ablation.pdf}
%     \caption{Comparison of tracking performance with and without the CLF decreasing condition in Mujoco. The average velocity for each part of the trajectory is shown in the dashed lines. The CLF-based tracking reward performs better than the pure tracking reward at higher speeds, highlighting the performance benefits of a CLF based reward.}
%     \label{fig:clf_ablation}
%     \vspace{-3mm}
% \end{figure}

Lastly, there are a few regularization rewards: a torque minimizing reward, an action rate reward, a reward to mitigate joint over-extensions, and a reward to mitigate torques over the limit of the motors. We can write these rewards as
\begin{align*}
r_{\mathrm{reg}} =\; &
- w_\tau \|\tau_t\|^2 
- w_{\Delta a} \|a_t - a_{t-1}\|^2 \notag \\
& - w_{q}\left\| 
\max\left(0, q_{\min} - q_t\right) + 
\max\left(0, q_t - q_{\max}\right) 
\right\|_1 \\
&- w_{\tau\text{lim}}\max(\||\tau_t| - \tau_{\text{lim}} \|_1, 0).
\end{align*}
In total the reward is therefore
\begin{equation}
    R = r_{\mathrm{reg}} + r_{v} + r_{\dot{v}} + r_{\mathrm{dom}}.
\end{equation}
The reward weights are given in Table \ref{tab:reward_table}.

\begin{table}
\centering
\caption{Reward weight coefficients used during training.}
\label{tab:reward_table}
\begin{tabular}{ll}
\toprule
\textbf{Reward Term}              & \textbf{Weight} \\
\midrule
Torque penalty \( w_\tau \)       & $1\times10^{-5}$ \\
Action-rate penalty \( w_{\Delta a} \) & $1\times10^{-3}$ \\
Torque limit penalty \(w_{\tau \text{lim}}\) &  $1.0$ \\
Joint limit penalty \(w_{q_{\mathrm{limit}}}\) & $1.0$\\
CLF tracking reward \( w_v \)     & $10.0$ \\
CLF decay penalty \( w_{\dot{v}} \) & $1.0$ \\
Holonomic position reward \( w_{\text{hpos}} \) & $4.0$ \\
Holonomic velocity reward \( w_{\text{hvel}} \) & $2.0$ \\
Contact penalty \(w_{\text{con}}\) & $3.0$ \\
\bottomrule
\end{tabular}
\vspace{-3mm}
\end{table}

\begin{figure*}
    \centering
    \includegraphics[width=1.0\linewidth]{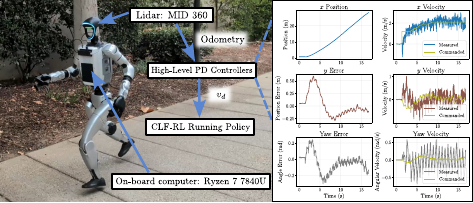}
    \caption{Experimental setup and outdoor velocity and position tracking. The graphic on the left shows the location of the on-board computer and lidar as well as the layered architecture for straight running. On the right we can see the hardware position and velocity tracking data from outdoor running. The policy tracks desired positions and velocities very well.}
    \label{fig:expermental_setup}
    \vspace{-3mm}
\end{figure*}

To achieve straight-line running, the robot must be able to correct errors in the yaw angle and in lateral drift. This requires the ability to respond to angular velocity commands about the $z$-axis as well as lateral velocity commands. Because the nominal reference trajectories do not contain these motions, they are modified during training when such commands are applied by adjusting the stance and swing foot locations and orientations, along with the center of mass position. The policy is trained on both isolated lateral and yaw velocity commands, as well as their combined use within a PD controller for straight-line running, which helps mitigate distribution shift when deploying with the PD controller on hardware.

Both the actor and critic networks use a fully connected feedforward architecture with three hidden layers, using the ELU (Exponential Linear Unit) activation function at each layer. The actor receives the base angular velocity, projected gravity, commanded velocity, joint angles, joint velocities, the previous action, and $\sin$ and $\cos$ phases as observations. The phase observations have a period of exactly one full gait period. To facilitate learning, the critic receives additional privileged information not accessible to the actor, such as stance and swing foot linear and angular velocities, reference trajectory positions and velocities, and binary contact state indicators. These inputs help stabilize value estimation and improve training efficiency. We use IsaacLab and IsaacSim for GPU-accelerated physics simulation~\cite{mittal2023orbit}. For policy training, we use the PPO implementation from Robotics System Lab RL library~\cite{rudin2022learning}. We also use the symmetry loss proposed in \cite{mittal_symmetry_2024}.

To facilitate sim-to-real transfer, we train the policy in simulation with domain randomization, a common practice that helps prevent overfitting to the simulator and improves robustness by exposing the policy to a wide variety of states that could arise from model mismatches in the real world. Specifically, we randomize the torso’s center of mass (COM) position, apply additional torso mass, modify PD gains, vary restitution coefficients, joint frictions, armatures, and ground friction, and apply periodic velocity impulses during training.

To incorporate the gait library during training, we use the commanded forward body velocity $v^d_x$ to select the closest speed gait from the library. The bezier curves defining the virtual constraints are then used to construct the CLF-based reward. The commanded forward velocity, $v^d_x$, ranges from 1.1 m/s to 3 m/s and 2\% of the environments are commanded to stand. For the lateral and yaw velocities, we command $v^d_y \in [-0.5, 0.5]$ and $v^d_{\omega} \in [-0.5, 0.5]$.

\subsection{Straight Running}
To enable running on a relatively narrow treadmill, the robot must avoid excessive lateral drift. To address this, we employ a layered control architecture in which independent PD controllers regulate both lateral motion and heading. The outputs of these PD controllers generate commanded velocities, which are incorporated into the RL observations.

To obtain the position and yaw of the robot, we use Fast-LIO \cite{xu_fast-lio_2021} which generates odometry data from the lidar. The velocity estimates are smoothed using a moving average filter and then provided to the derivative term of the controller, while the position estimates from Fast-LIO are used directly without modification. Finally, the controller outputs are saturated according to the training limits of the commanded $v_y$ and $v_\omega$ velocities.

% \subfile{sections/implementation}

\section{Results}
\label{sec:results}

\begin{figure*}
    \centering
    \includegraphics[width=1.0\linewidth]{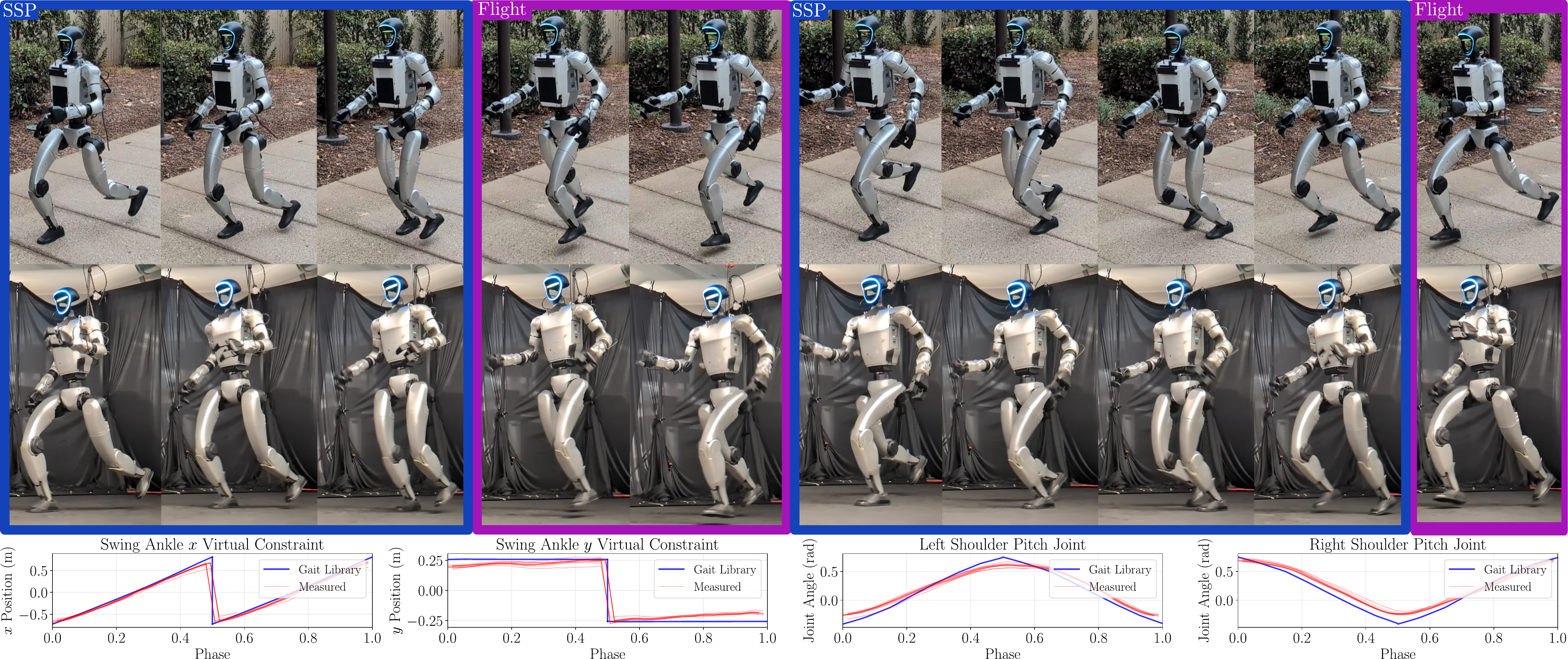}
    \caption{A running gait tile showcases one full period inside and outside. The single support phase (SSP) and flight phase are highlighted here to emphasize the multi-domain behavior realized on the physical system. Further, one can see how repeatable the gait is both inside and outside by comparing the running motions in each location. Select virtual constraint tracking plots from outdoor hardware data are shown below. Note that the swing ankle values are computed only approximately using forward kinematics. One can see that the policy is able to track well, especially in the $x$ direction swing foot.}
    \label{fig:gait_tile}
    \vspace{-3mm}
\end{figure*}

We evaluate the proposed controller both in simulation and through extensive hardware experiments. In particular, we investigate the effect of the dynamics in the optimization, the policy’s trajectory tracking performance, its global velocity and position tracking accuracy, as well as its overall performance and robustness on hardware. The learned policy is deployed on a Unitree G1 humanoid, with sensing and compute contained entirely onboard. Indoor treadmill experiments demonstrate the robot’s ability to run along a confined path and recover from external disturbances, while outdoor trials show the policy’s effectiveness in real-world environments.

First, we study the effects of using dynamics information in the optimization in comparison with a kinematic optimization. It is easier to generate kinematic trajectories, but we demonstrate that for motions like running the dynamics information is critical. We run the same trajectory optimization but swap the dynamics constraint for a kinematic constraint, i.e. enforcing the integration of velocity to position while ignoring forces and accelerations. The results are shown in Fig. \ref{fig:kinematic_comparison} which demonstrate that the dynamic reference trajectories consistently out-perform the kinematic trajectories both in training metrics but also in reference tracking. We run each training with the same 5 unique seeds to show that this comparison, and the RL training, is consistent across seed. The tracking error improves by about 1.7x over the kinematic reference which is a substantial improvement. Note that for this comparison we choose to track a single speed rather than the whole range to isolate the difference in an isolated reference tracking paradigm. 

Secondly, we examine the trajectory tracking performance of the policy in simulation. Fig. \ref{fig:isaac-traj-tracking} demonstrates how the policy can track the virtual constraints. Six out of 27 total virtual constraints are shown in the plot; we focus on the center of mass (COM) position and some states of the swing ankle. Although no transient motions are given in the gait library, the policy is still able to approach the desired orbit well which can be seen in the first step period on the plot. In general the policy has good trajectory tracking performance. Note the trajectory optimization has no notion of robustness and can use the full dynamic capability of the robot while the RL must learn to make a stabilizing policy even with domain randomization. For this reason we cannot expect the policy to track the trajectory perfectly, and this explains why the COM $z$ tracking is not as good as the others - the policy sacrificed some of the tracking for robustness.

The policy demonstrates strong performance in outdoor environments using only on-board compute and sensors, as depicted in Fig. \ref{fig:expermental_setup}. The onboard computer is used for policy inference, while the lidar module's odometry is used for position and velocity tracking. The $y$ (transverse) position and yaw are driven to zero using PD control, while the $x$ (sagittal) direction tracks the commanded velocity. The plot shows that the policy is able to accurately track the velocity command, even outdoors, while keeping the magnitude of the $y$ error less than 0.25 m and the yaw error less than 0.2 rad, after the transient. This highlights the tracking capabilities of this policy and underscores its ability to be placed into a full autonomy stack.

Figure \ref{fig:gait_tile} illustrates the motion of a running stride performed both indoors and outdoors. The gait tile highlights when the robot is in single-support versus flight phases, showcasing the multi-domain structure of the running gait. These results demonstrate that the robot can reliably run in both environments. The figure also includes selected virtual constraint tracking plots, where the first half of the phase corresponds to the left foot in swing and the second half corresponds to the right foot in swing.

\begin{figure}
    \centering
    \includegraphics[width=1.0\linewidth]{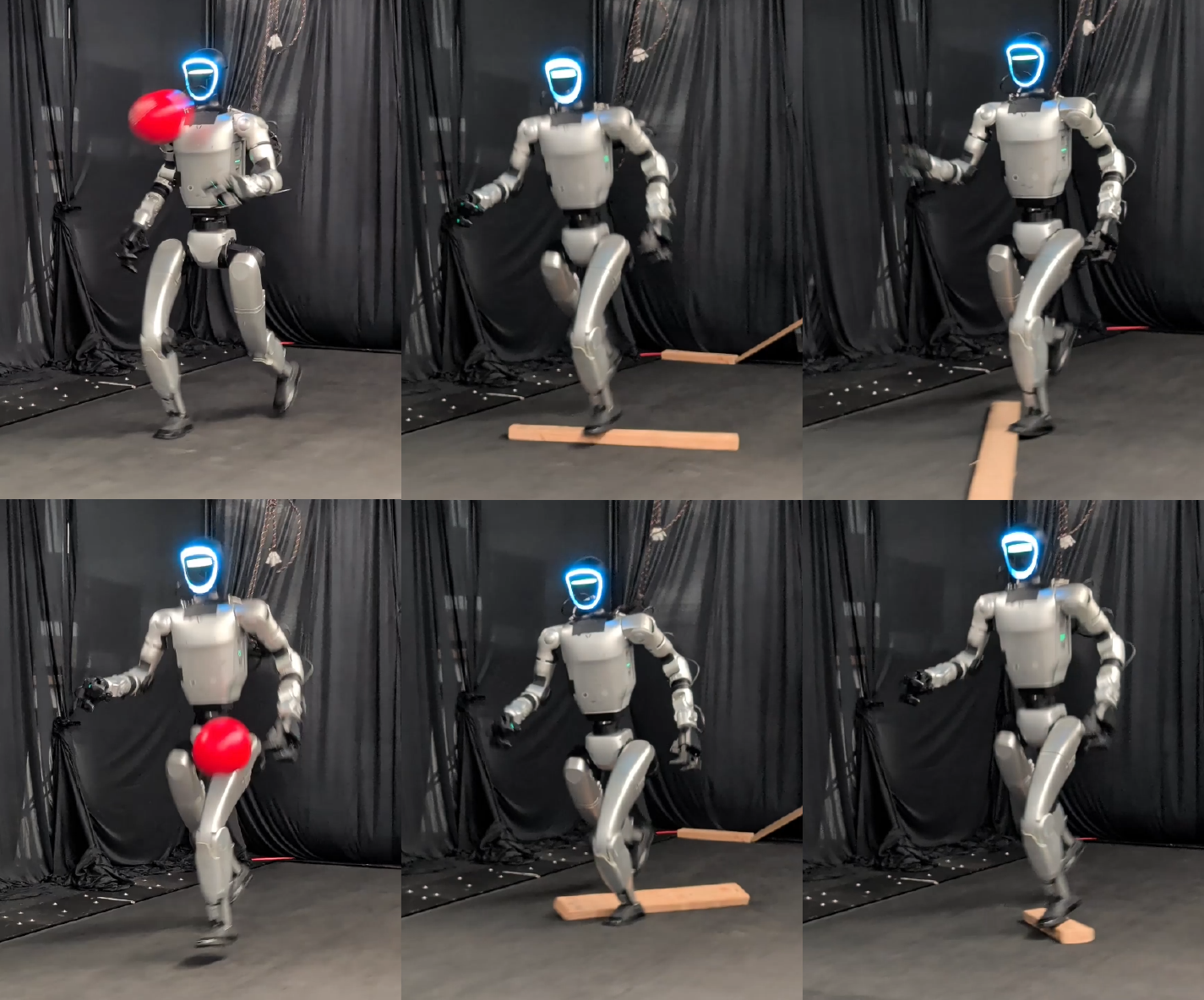}
    \caption{Robustness testing with the running policy on the treadmill. The policy is able to handle dodge balls being thrown at it and wooden planks of various sizes hitting the feet or being landed on. The robot successfully recovered from every disturbance pictured here. No object detection or avoidance was used; only the running policy.}
    \vspace{-4mm}
    \label{fig:experiments}
\end{figure}

Beyond demonstrating basic running, we evaluate the policy through a series of robustness tests. While running at approximately 2.2 m/s on the treadmill, the robot is subjected to external disturbances, including being hit with dodge balls and having wooden planks sent down the treadmill towards its feet.  Fig. \ref{fig:experiments} shows six such tests that the robot was able to recover from successfully. These include impacts from dodge balls to the torso and leg, wooden planks striking the stance toe and heel, and the swing toe kicking a moving plank. These experiments demonstrate that the running policy is robust and can withstand various disturbances while maintaining high-speed locomotion with a flight phase.

% \begin{figure}
%     \centering
%     \includegraphics[width=1.0\linewidth]{sections/paper_tracking_performance_outdoor_9_12_25_v1.pdf}
%     \caption{Outdoor velocity and position tracking.}
%     \label{fig:outdoor_tracking}
% \end{figure}

\section{Conclusion}
\label{sec:conclusion}
We presented a complete framework for generating robust and high-performance policies for humanoid running. Full-order dynamic reference trajectories are generated offline using a multi-domain trajectory optimizer. Based on this gait library, a control Lyapunov function (CLF) tracking controller is synthesized and embedded into the RL reward, replacing heuristic reward design with a dynamically feasible trajectory and stability informed reward. The policy is trained in simulation and deployed on a Unitree G1, where it produces natural running with a full flight phase, accurate position and velocity tracking, and robustness to disturbances using only onboard compute and sensing. Furthermore, this framework is readily extendable to other motions, such as jumping or cartwheels, and allows them to be highly optimized for any given robot, pushing the boundary of dynamic motions for humanoid robots.
% The policy is then trained in simulation and subsequently deployed on hardware. 

%The resulting policy produces natural and highly dynamic running behaviors with a full flight phase on a humanoid robot. Beyond its dynamic capabilities, the controller exhibits accurate position and velocity tracking. The running is robust to disturbances and functions both inside on a treadmill and outside in every-day environments, using only compute and sensing contained onboard. Furthermore, this framework is readily extendable to other motions and allows them to be highly optimized for any given robot, pushing the boundary of dynamic motions for humanoid robots.

\bibliographystyle{IEEEtran}
% \balance
\bibliography{IEEEabrv, References}

\end{document}